%% file: main.tex
\documentclass[pmlr]{jmlr}

\usepackage{multicol}
\usepackage{xcolor,colortbl}
\usepackage{multirow}
\RequirePackage{graphicx}
 \usepackage{booktabs}
\usepackage{longtable}
\makeatletter
\def\set@curr@file#1{\def\@curr@file{#1}} 
\makeatother
\usepackage[load-configurations=version-1]{siunitx} 

\theorembodyfont{\upshape}
\theoremheaderfont{\scshape}
\theorempostheader{:}
\theoremsep{\newline}

\title[Clinicians' Voice: Fundamental Considerations for XAI in Healthcare]{Clinicians' Voice: Fundamental Considerations for XAI in Healthcare}

\author{\Name{Tabea E. R\"ober}
       \Email{t.e.rober@uva.nl}\\ 
       \Name{Rob Goedhart}
       \Email{r.goedhart2@uva.nl}\\ 
       \Name{\c S. \.Ilker Birbil}
       \Email{s.i.birbil@uva.nl}\\
       \addr Amsterdam Business School, University of Amsterdam, The Netherlands        
       }

\begin{document}

\maketitle

\begin{abstract}
Explainable AI (XAI) holds the promise of advancing the implementation and adoption of AI-based tools in practice, especially in high-stakes environments like healthcare. However, most of the current research lacks input from end users, and therefore their practical value is limited. To address this, we conducted semi-structured interviews with clinicians to discuss their thoughts, hopes, and concerns. Clinicians from our sample generally think positively about developing AI-based tools for clinical practice, but they have concerns about how these will fit into their workflow and how it will impact clinician-patient relations. We further identify training of clinicians on AI as a crucial factor for the success of AI in healthcare and highlight aspects clinicians are looking for in (X)AI-based tools. In contrast to other studies, we take on a holistic and exploratory perspective to identify general requirements for (X)AI products for healthcare before moving on to testing specific tools. 
\end{abstract}

\section{Introduction}
\label{sec:intro}
Rapid developments in artificial intelligence (AI) call for more transparency of AI-based technologies to tackle concerns related to trustworthiness, responsibility, and accountability. This has led to the emergence of the field of \textit{Explainable AI} (XAI). The goal of XAI is making AI models more transparent and understandable to address these concerns and aid the implementation of AI-based technologies. Especially in high-stakes situations, like  those in healthcare, AI-based systems could enhance critical decision-making processes \citep[\textit{e.g.,}][]{Eke2024,Alowais2023}. Ideally, AI models provide a certain prediction (\textit{e.g.}, the probability of a skin lesion being cancerous) which supports the clinicians in their decision process. However, implementation in the medical field -- beyond regulatory requirements these technologies have to meet -- has been slow partly due to lacking transparency \citep{Eke2024,He2019}.
Although user studies, evaluations, and multidisciplinary research are generally advocated for within XAI, only a small fraction of publications evaluate their proposed solutions with users. For example, \cite{Saarela2024} reviewed 512 articles, where almost 60\% did not do any evaluation at all, and only 8\% performed qualitative evaluation involving domain experts. Similarly, \cite{Nauta2023} report that only 20\% of over 300 papers they reviewed performed evaluations with users. 
Arguably, XAI approaches to medical imaging as used in radiology may be most frequently implemented, as shown by the number of publications with XAI applications in this sub-field alone \citep[\textit{e.g.,}][]{Borys2023, Gupta2024}. 
For tabular data, however, there are less XAI applications, implementations and user tests. Therefore, our work focuses on tabular data structures like electronic health records.
Specifically, in this research, we conduct semi-structured interviews with clinicians from the Netherlands. Our aim is to elicit clinicians' thoughts on XAI and, by that, uncover compelling perspectives that may give rise to new research avenues and considerations in the design of XAI methods for healthcare and especially for tabular data structures. For example, feature importance scores are often praised for their simplicity, however our research suggests that they may become unbelievable due to exactly this property. Furthermore, our interviewees also raised concerns about taking responsibility for decisions, even if there is sufficient trust in the AI.

\subsection*{Generalizable Insights about Machine Learning in the Context of Healthcare}
Overall, our interview study provides compelling statements and unique angles that put existing research into a new light and could inspire new research. We also find support for previous research, including that users emphasize the importance of receiving information about the background of the model in addition to its predictions and explanations. Our results further highlight the importance of multidisciplinary projects to ensure seamless introduction of (X)AI-based tools into clinical practice, and the significance of training clinicians on AI before implementing AI-based tools in their workflow. 

Studies like this one are often overlooked in machine learning research albeit their importance for developing holistic AI solutions for healthcare. Generalizing findings and claims about interpretability of AI models requires keeping clinicians in the loop, and multidisciplinary collaboration at all stages of the design and development process is key to successful implementation and bringing AI to the bedside.



\section{Background}
\label{sec:background}

One can generally distinguish between \textit{glass-box} and \textit{black-box} models in XAI. Glass-box models are transparent, and model internals and operations are accessible to the users so that they can trace back every prediction made by the model. Thus, glass-box models provide both local and global explainability. Black-box models are opaque machine learning models where the user is unable to trace back the prediction due to the model complexity. Typically, (only) local explainability is achieved by complementing single predictions made by black-box models with \textit{post-hoc} explanations. For an in-depth account of developments in XAI, we refer to several recent review papers \citep[see, for example,][]{das2020_review, gilpin2018_review, murdoch2019_review, ras2022explainable, linardatos2021_review, guidotti2018_review, carvalho2019_review, barredo-arrieta2020_review, ali2023_review}

Healthcare is a high-stakes domain with vast opportunities for successful (X)AI implementations. This is also reflected in the increasing number of publications on AI-based clinical decision support tools (DSTs) that incorporate explainability. For example, \citet{Thoral2021} trained an AI-model to predict readmission or death within seven days after discharge from the ICU and explain predictions using feature importance scores \citep{Lundberg-SHAP}. \citet{abdullah2021_review} and \citet{antoniadi2021_review} provide reviews of explainable machine learning approaches in healthcare, and \citet{markus2021role} provide a step-by-step guide to choose the most appropriate XAI method.
\citet{amann2020explainability} take a multidisciplinary approach to establish the need for explainability in healthcare from technological, legal, medical, patient, and ethical perspectives. 

While it is generally expected that complementing AI systems with explanations will increase task performance and trust \citep[\textit{e.g.,}][]{Senoner2024, lim2009}, there is also research suggesting that explanations complementing AI outputs may lead to over-reliance. For example, \cite{VANDERWAA2021103404} and \cite{jabbour2023measuring} show that AI-based DSTs with explanations do not help participants identify incorrect predictions.
Alarmingly, \cite{Kaur-interpretability-2020} showed that even data scientists tend to over-rely on AI systems if explanations are provided. These results emphasize the need for a careful design process considering contextual requirements and the interplay of humans and technology.

\citet{Cai2019} conduct a user study including semi-structured interviews on AI-assisted diagnosis of prostate cancer, focusing on identifying desired aspects for explainability in the AI tool. 
They find that pathologists are interested in global properties of the AI tool beyond accuracy, such as how conservative or liberal its decisions are or what kind of data it was trained on. In a similar fashion, \citet{NAISEH2023102941} empirically evaluate the effect of different types of explanations on trust calibration in a medical context and also conduct interviews with a smaller group of participants. 
Like \citet{Cai2019}, they find that practitioners are interested in global information of AI-tools, such as the contents of the data and the recency of the trained model. They further find that practitioners worry about fitting AI-tools into their workflow because it adds extra workload and that they would like more guidance on how to interpret the explanations.
\citet{pmlr-v106-tonekaboni19a} conduct interviews with 10 clinicians that are familiar with AI-based clinical tools, primarily to identify which explanation types best serve their needs and to propose evaluation metrics of explainability tools. For example, they find that clinicians value feature importance scores as they draw attention to specific patient characteristics relevant for the prediction. They further conclude that example-based explanations are less helpful as they depend on the definition of similarity with the added difficulty that seemingly similar patients may have significantly different outcomes and vice versa. They also find that accuracy alone is not sufficient and that the output should be complemented with a certainty score for the prediction.
Other qualitative studies address, for example, onboarding material for XAI-tools \citep{lee2024improvinghealthprofessionalsonboarding}, adopt a broader stakeholder perspective \citep{SUBRAMANIAN2024102780}, or analyze collaboration between developers and clinicians \citep{Bienefeld2023}. 


Our study differs from this previous research in several ways. First, we do not build our work around a specific case or AI-tool. 
We aim to elicit and analyze perspectives of clinicians that may be universal to a range of AI products and fields within healthcare.
Second, we interview clinicians from different fields within healthcare to record general data. Third, our study is exploratory in nature rather than empirically driven. That is, we do not test for an effect of XAI methods on, for example, trust or understanding.  
Qualitative interview studies are known to be exploratory in nature and generalizing findings may be difficult. Nonetheless, they pose a useful tool to gather in-depth understanding that can enrich results of quantitative studies and to uncover unique perspectives and angles for further research \citep{qualitative-study, Dunwoodie2023}. 


\section{Methodology}
\label{sec:methods}



We conducted 11 semi-structured interviews with open-ended questions with clinicians trained and working in the Netherlands. In this section we discuss recruitment strategy, interview procedure, materials, and limitations of our design. The supplementary materials for the interviews are available in the appendix. The study was approved by the ethical body of our institution, the Economics and Business Ethics Committee of the University of Amsterdam, The Netherlands. Interviewees were informed of the goal and procedure of the interviews and signed an informed consent form prior to their participation. 

\subsection{Participants and recruitment}
The sample consisted of six women and five men, with years of experience in their fields ranging from two to 30 years. The main inclusion criterion for participants was successful completion of their medical study in the Netherlands. The sample included clinicians from different fields within medicine (general practice, (pediatric) oncology, radiology, general health psychology, obstetrics and gynecology, surgery). Participants were recruited through networking and word of mouth. None of the interviewees was personally known to the interviewer before the interview was conducted. 

\subsection{Procedure}

Potential interviewees were contacted via email, which stated the reason for reaching out, the person that had provided the contact details, practical information such as the expected time, and the informed consent form. If interviewees responded positively, an exchange followed to arrange a meeting. Interviews were scheduled for 30 minutes, and the majority of them took place online. 
During the interview, the interviewer confirmed with participants that they agree to have the interview audio-recorded. Interviewees could ask questions or withdraw from the interview at any moment, and were explicitly given the opportunity to add any information at the end of the interview. All interviews were conducted by the same interviewer to avoid potential differences due to interviewing style.

\subsection{Materials}

An interview scheme with guiding questions was developed to provide structure and guidance to the interviews (see Table \ref{tab:interview}). Broadly, aspects that we looked at include (i) tasks that clinicians (would) find (X)AI helpful for, (ii) desirable characteristics of XAI methods/explanations, (iii) hopes, concerns, and fears they face regarding the implementation of (X)AI in their workflow, and (iv) their perception of the role of trust in the implementation and adoption of (X)AI.

\input{tables/interview-scheme}

At the start of each interview, the interviewer provided context for the interview after introductions, similar to what participants had read in the informed consent form beforehand. Then, the first set of questions was aimed at inquiring about the interviewee's background, including their specialization and experience. The following questions were aimed at understanding the current use of AI in their field and healthcare in general. Besides, we also asked interviewees to share about research in (X)AI for healthcare that they are aware of. Then, we encouraged them to come up with use cases of AI in their field which they would find helpful in their job. Following up, we asked them for reasons why these do not yet exist, and why implementation of AI in healthcare in general is slow. The next set of questions was focused on uncovering interviewee's attitude towards AI in healthcare. This entailed asking specifically for their positive and negative feelings towards using AI in healthcare, and if they could describe their hopes and concerns. In this part of the interview, we also asked them about the role they believe trust plays in the implementation and adoption of (X)AI in practice. Then, for the next set of questions, the interviewer explained three different types of explanations using a simple illustrative example for illustration (see Appendix \ref{app:interview_material} for the exact material and information given to participants) and asked interviewees to explain their thoughts considering each of these explanation types. The explanations that we chose were (1) feature importance scores, (2) counterfactual explanations, and (3) a glass-box model. 

\paragraph{Feature importance scores.} Feature importance methods calculate a score for each input feature that shows its contribution to the predicted score.
We decided to include this method as it is arguably the most popular method within XAI at this moment. For our illustrative example, we utilized scores and the respective bar plot produced by SHAP \citep{Lundberg-SHAP}. Starting from the mean prediction in the data, a positive SHAP value increases the score while a negative value decreases the score. Due to the additive nature of SHAP, adding up the SHAP values of all features is equivalent to the final prediction.

\paragraph{Counterfactual explanations.} Counterfactual explanations show the minimal change required in the input feature space to change the prediction of a sample to a different class. In the literature, the sample and its corresponding prediction that is to be explained is coined \textit{factual}, whereas the changed data point leading to the desired prediction is called \textit{counterfactual}.  
Research from the social and cognitive sciences shows that counterfactual explanations mimic the way humans give explanations, and hence are expected to be good explanations in XAI \citep{byrne-ijcai2019p876, MILLER20191}. This has spurred many researchers to propose algorithms generating these explanations \citep[\textit{e.g.,}][]{Russell.2019, Ustun.2019, Kanamori.2020, Kanamori.2020jn, Karimi.2019fy, Karimi.2020, Mothilal.2020, Mahajan.2019, maragno2022counterfactual}. 

\paragraph{Rules as glass-box model.} As a contrast to the previous two explanation methods for black-box models, we opted for a glass-box model where the model itself is transparent and predictions can be traced back. 
For our illustrative example, we chose the method proposed by \cite{rober2024rulegenerationclassificationscalability}, which produces a set of short and weighted rules. Each sample is classified by a subset of the rules, and rules receive weights which can be interpreted as importance. The final prediction stems from a weighted combination of the rule predictions. 

Interviewees received a concise but detailed information sheet as supplementary material before the interview, which included the illustrative example and the three explanations, complemented with information about what they show and how to interpret them. We include the material that interviewees received in Appendix \ref{app:interview_material}. 
Besides their opinion on these specific explanation types, we included a question on what characteristics (\textit{e.g.}, visual, interactive) they would like to see in an AI system, or what type of information they would like to be shown. Then, at the end of the interview, interviewees were asked if they would like to share any other considerations with us. 

\subsection{Analysis}

Upon completion of all interviews, each recording was transcribed and subsequently analyzed using inductive thematic analysis \citep{Braun01012006}. Inductive thematic analysis is a flexible methodology for analyzing qualitative data in a bottom-up manner, \textit{i.e.}, not driven by existing theory. In this data-driven approach, interviews are coded systematically by identifying features relevant for each aspect of the interview scheme and the overall research goal, collated by data relevant to each code. This step helps to compress the data into digestible chunks that at the same time retain the relevant information. Subsequently, themes are identified by grouping codes together, by that forming broader patterns of meaning related to the research goal. In this bottom-up and exploratory analysis, no predefined coding categories or themes are used; instead, responses are carefully analyzed to identify recurring themes and patterns and the codes are developed organically in response to the data. 

\subsection{Limitations}

The nature of our study bears some limitations we should consider when interpreting the results. First, interviews are known to bear threats to validity and reliability \citep{alshenqeeti2014interviewing}. As a response to this, we chose to have the same interviewer conduct all interviews. Further, interviews were audio-recorded and then transcribed for the analysis, by that avoiding misinformation through note-taking or incorrect understanding by the interviewer. Furthermore, during the interviews, time was taken to clarify any unclarities to avoid misunderstanding and misinterpretation of the responses. Second, as an explorative interview study, the sample size is limited and we did not perform any quantitative analyses of the results and should be careful with making generalizations. Third, the interviews were scheduled to last 30 minutes, which may not be considered long enough to discuss different aspects of XAI. As this is an exploratory study and clinicians are known to have limited time, we were constrained by this duration. Lastly, we reached out to candidates via email, and  in some cases a common connection made the introduction. Generally, it can be expected that individuals interested in XAI are more likely to positively respond to an interview request. Hence, we may only have interviewed individuals with strong (positive or negative) opinions on the topic.

\section{Results}
\label{sec:results}

We next discuss patterns identified through our thematic analysis of the interviews and put them into perspective using direct quotes and examples from interviewees. While reserving our main discussion for Section \ref{sec:discussion}, we discuss in this section the relevant literature directly linking to interviewee's responses. 

\subsection{Current use of AI in healthcare}

Interviewees generally talked about AI-based tools in healthcare in three different categories, namely tools related to (1) business operations, (2) care processes, and (3) medical device regulations.
The first category of AI systems is relatively universal to any larger business, not only hospitals or care facilities, and covers aspects like finance, forecasting, and so on. The second category relates to AI systems that enhance the care process by, for instance, making it more efficient (\textit{e.g.}, the planning of chemotherapy).
The last category of AI systems in healthcare concerns AI systems that fall under the medical device regulation. Simply put, a medical device is any instrument intended to be used to diagnose, treat, monitor, or prevent diseases ``and does not achieve its primary intended action by pharmacological, immunological, or metabolic means'' \citep[p.7]{IMDRF-2013}. AI models fall under the category \textit{Software as a Medical Device} and are commonly referred to as \textit{decision support tools}, meaning that the AI system's output is intended to support clinicians in their decision process.

When asked to what extent AI models are implemented in medical care in their field specifically, interviewees unanimously said that there is very little AI (as DSTs) integrated in their workflows. Most interviewees (7/10)\footnote{Due to the nature of semi-structured interviews not every question was posed to each interviewee. Therefore, the number of respondents may differ per question and we indicate it accordingly.} pointed out that a significant amount of budget is invested in research and development of medical AI models. Although many of these efforts are being tested, very few of them actually reach the bedside. Arguably, the most popular field for AI applications within healthcare is radiology, with the number of publications on AI in radiology being ``sky high'' (interview 4). However, even though there are more than 200 products in AI for radiology available on the European market\footnote{\href{https://radiology.healthairegister.com}{https://radiology.healthairegister.com}}, not many of those are integrated in clinical practice yet. Furthermore, one interviewee pointed out that most of these products perform specific subtasks, such as segmentation or quantification tasks, instead of predicting a concrete medical outcome; thus, only indirectly supporting clinicians.

However, some interviewees (3/9) also mentioned that other forms of AI models, such as those for people planning and scheduling, are integrated everywhere, and considering AI in that sense ``everybody is really already using AI a lot'', but ``not [...] as a medical device yet'' (interview 10). Note that this falls into the second category of AI models previously mentioned.

\subsection{What tasks should AI be used for?}

Even though some interviewees held the opinion that AI for administrative tasks is already widely used, others (4/9) said that they would want to see AI be more used for administrative and logistics tasks. One interviewee said that ``80\% of our task is administration'' (interview 6), hence if AI could be used for this type of work, clinicians could focus more on their patients. Two interviewees mentioned a type of large language model for medical purposes, for example to help them write summary reports which is a tedious task taking up a significant amount of time.

Regarding AI as a clinical DST, four out of nine interviewees mentioned predicting complications as a useful application. For example, when performing surgery, they would like to know upfront how large the chance is to run into a certain complication. And even beyond that, one interviewee said they ``would like to know upfront what can go wrong and how can we prevent that'' (interview 5). Hence, AI could serve not only as a warning system but also as a guide on how to navigate certain situations. To give a concrete example, one interviewee from obstetrics and gynecology explained that when a woman is in labor, they monitor both the mother and the baby, meaning they collect a large dataset. Having an AI model trained on this data that continuously predicts the chance of a birth ending in a C-section would save them not only time and resources, but would also reduce the risks for women in labor. Another example comes from oncology, where it would be beneficial to predict the likelihood of complications during surgery but also the probability of remission following treatment.

Two out of nine interviewees mentioned that monitoring or medical surveillance would be helpful, especially in cases where patients are not continuously at the hospital. With the recent technological advancements, it may be possible to provide a wide range of patients with wearables to monitor their health. In this case, (X)AI-based tools could be used to predict time of admission to the hospital or make recommendations on further treatment, which could then be evaluated in consultations and check-ups with the clinician.

\subsection{Attitude towards (X)AI in healthcare}

\paragraph{Positive attitude.}

The most common response (5/8) was that AI is simply more efficient than humans and can ``make our life easier'' (interview 3). Again, it was highlighted that AI can take over largely the tedious tasks, hence acting not necessarily as a DST but rather taking over relatively well-defined but time-consuming tasks, such as planning and writing reports.

All interviewees had mainly positive thoughts about AI-based DSTs, most of them (6/8) believing that AI can help them make the best decision. As a human being, ``it's really difficult to make complex decisions and to take into account more than four or five variables'' (interview 9).
However, it is not only about how many variables clinicians need to take into account when assessing the patient's situation, but also relating these to their domain knowledge and previous cases. For example, one interviewee pointed out that AI ``has a memory of a lot more cases than I can experience and remember'' (interview 2), so it can base its prediction on a much wider range of previous patients. Another was convinced that ``you can treat a patient better with the data you have'' (interview 5), which will ``make it better'' (interview 5) for both the patient and the doctor. 

While interviewees seemed to agree that AI can support them in the process of making the best decision for their patients, two (of eight) explicitly mentioned that AI could do so by taking a more objective stance. That is, sometimes our judgments can be clouded by emotions and especially in fields where there is frequent clinician-patient contact one may have too much of a ``tunnel vision on [a] patient'' (interview 3), in which cases it would be helpful to have ``an objective somebody next to you'' (interview 3). \\

\noindent
\textit{Link to literature.} Research in the cognitive sciences has indeed shown that humans can only consider a limited amount of information at a time \citep[\textit{e.g.,}][]{miller1956magical, Cowan_2001, Luck1997, OBERAUER2006601}. Working memory capacity has also been linked to decision-making, and humans tend to rely on heuristics to make decisions if cognitive demands are too high \cite[\textit{e.g.,}][]{broder2003take, shah2008heuristics}.

\paragraph{Negative attitude.}
However, interviewees (2/8) also voiced concerns that the distance between them and their patients will become larger when using AI-based DSTs. The same case may have different outcomes because of the patients' wishes, beliefs, and preferences -- ``soft features'' (interview 2) that are not captured in the data. Personalized treatment is only possible if you interact and connect with your patient, know their story, and incorporate their wishes. Using AI-based devices for some tasks bears the fear that the distance between doctors and their patients becomes so large that they will be dependent on these devices. This outcome would take away a large part of what drives many clinicians, which is simply the interaction with their patients. However, it is important to note that four other interviewees, when asked about their opinion on this aspect, said that they are not concerned about it because regardless of where any medical information comes from, it is still them (the clinicians) who are in charge of the situation and in control of the decision. As the name suggests, a DST is merely there to give support -- the decision still lies with the clinician who takes them in consultation with the patient.

Bias and AI literacy are other aspects that concern clinicians. In the context of AI, bias relates to the data the model was trained on, and has been an important topic in recent research, especially after criticisms were voiced against some algorithms implemented in practice\footnote{An example is an AI-based recruitment tool in testing at Amazon that discriminated on the basis of sex  \cite{bbc-2018}}.
Bias in AI exists when the model favors certain people, that is an AI-based DST may provide more accurate guidance for a certain group of patients. For example, if the dataset used for training included a higher proportion of elderly people, the model might give misleading or incorrect results for younger patients. If such systems were to be used, clinicians should be aware of these drawbacks to accurately interpret the output given by the device. Several interviewees pointed this out and referred to it as \textit{validity}. In fact, five (out of nine) interviewees mentioned that the validation of AI-based devices is extremely important and difficult, making it hard to develop models that can be applied to all patients. Overall, interviewees in this study showed that they are aware of the issue, and further pointed out that this is especially concerning if clinicians rely too much on the AI model instead of using it complementary to their expert knowledge. This concern was voiced especially for younger generations of clinicians who are and will be developing their expertise while these systems are around, potentially making them less critical towards them. 

Generally, participants seem to agree that AI is the future and can solve many problems. However, is this really true? To quote one of the interviewees, AI is ``kind of the Holy Grail to a lot of people'' (interview 6), without considering the context. In line with that, this participant also emphasized that we should always start with the simple models and only move to black-box models if necessary. \\

\noindent
\textit{Link to literature.}
Recently, advocates of glass-box models have highlighted that simple models like (optimal) decision trees and other rule based methods perform similarly if not better to complex black-box models, especially in cases where the data is structured like in tabular data \citep{rudin2019_stopexplaining, Semenova2022_10.1145/3531146.3533232}. In fact, the well-known trade-off between performance and interpretability has also been coined as a `myth' \citep{rudin2019_stopexplaining}.  

\subsection{What's best -- feature importances, counterfactual explanations, or rule-based methods?}

Comparing only feature importance scores, counterfactual explanations and rule-based models, there is a clear tendency for people to prefer feature importance scores. Interviewees generally agreed that they are intuitive, quick and easy to understand, as you can grasp them all with one glance. It is so easy that ``you don't need to be a professional to understand what's going on here'' (interview 3). While this is clearly a pro of this method, it also raises concerns -- ``the [...] question is, is it really then that simple?'' (interview 5). Especially in complex decisions that require consideration of a multitude of factors, it is almost ``not believable'' (interview 1). Interviewees (2/10) also pointed out that they are familiar with this method of explaining the output of AI models, and hence may be biased in preferring it over the other two methods.

Counterfactual explanations require a bit more time to be understood, however they can provide one with more in-depth insights. As one interviewee put it, this type of explanation is more ``fluid'' (interview 5), rather than static as the feature importance scores, and it shows the effect of something changing. This property could also be used to improve prognosis, which however raises the question of causality -- a change in features leads to a different prediction by the model, which does not always translate to reality. Lastly, one interviewer explained that looking at counterfactuals means one is essentially looking at other patients, and not focusing on the current patient anymore. On top of that, there are endless other patients out there, so why would one look at this specific counterfactual? 


Rule-based methods are similar to how decisions are made --\textit{e.g.}, ``Of course, if you are a two year old, you will never have a mama carcinoma'', or breast cancer in layman's terms (interview 3)--, and they're helpful if you really want to understand the prediction. Knowing the decision mechanism makes the AI's prediction more ``believable'' (interview 1), gives a sense of control, and something concrete to base your decision on, which is especially helpful in high stakes decisions. 
One interviewee (out of 10) held the opinion that such simple model should always be preferred and that we should only move towards complex AI models (and with that other types of explanations) if the simple models are not enough. 
However, two of ten interviewees also pointed out that if you have many rules, this may be too complicated to use on the spot. For research purposes, when you want to uncover the mechanism behind patient characteristics and medical conditions or outcomes, rule-based methods are helpful -- but in clinical practice? 

Regardless of the type of explanation, visual explanations are preferred for being easily understood on the spot, and because it is a way to remove any potential language barriers. Another aspect interviewees raised was uncertainty of the model prediction -- being shown how certain the model is about a prediction makes it more trustworthy and helps clinicians navigate the usage of the information given. This ties in with knowing more about the background of the model, such as which data were used for training and testing, how it was validated, how big the sample was, and so on, which came up in about half (five) of the interviews without having posed a question targeted to this. \\

\noindent
\textit{Link to literature.} Statements by interviewees about counterfactual explanations, and specifically which counterfactual is shown, can be linked to discussions of distance measures in technical papers on counterfactual explanations. That is, the choice of distance function influences which counterfactual is found \citep{Guidotti2024}. Besides, most approaches use simple distance functions \citep[\textit{e.g.,}][]{maragno2022counterfactual, Maragno2024-robustCE, wachter2017counterfactual, Kanamori.2020}, however `closeness' in real situations may differ from mathematical definitions.

\subsection{The role of trust in AI}

About one third (3/8) of the interviewees mentioned trust comes automatically when a device is tested, certified, and approved. This is comparable to trusting lab results and equipment without exactly knowing how they work, or when they have last been calibrated. In line with that, trust develops over time as you use a device and experience its benefits. 
Especially more senior clinicians seem to be reluctant to use new AI-based medical devices. One interviewee raised the point that trust is distinct from responsibility; that is, you can trust a device but in the end you have to take responsibility for your decisions, and ``how are you going to feel okay with that responsibility when it's not your own thought?'' (interview 3). This adds a new angle to the discussion on trust of AI models that has not yet been sufficiently addressed in research. \\

\noindent
\textit{Link to literature.} It is not unusual that adoption of new products takes time and is not adopted at the same rate by everyone, so this aspect is not specific to AI in healthcare. In economics and sociology, the group of people that are hesitant towards new products are known as late majority and laggards \citep{rogers2003diffusion}.

\section{Discussion}\label{sec:discussion}

\begin{figure}[h]
    \centering
    \includegraphics[width=0.6\linewidth]{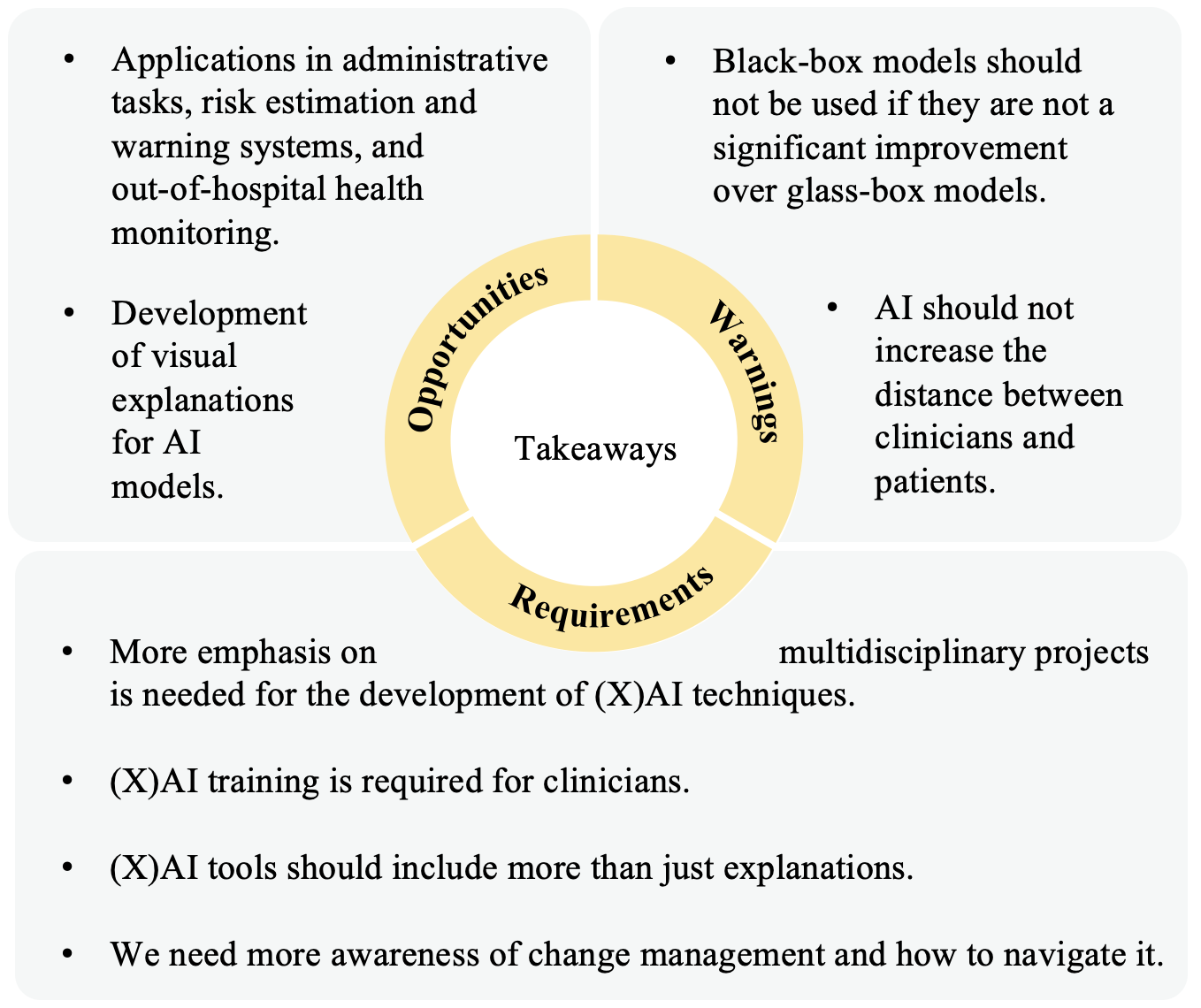}
    \caption{Overview of key takeaways formulated on the basis of the conducted interviews.}
    \label{fig:key-findings}
\end{figure}

Based on the results obtained, we highlight the opportunities for (X)AI research, the requirements for successful implementation of (X)AI in practice, and the main warnings provided regarding (X)AI usage and development; see Figure \ref{fig:key-findings} for an overview.

\subsection{Opportunities for (X)AI research}
\paragraph{Applications in administrative tasks, risk estimation and warning systems, and out-of-hospital health monitoring.} The interviews have shown a positive attitude towards further (X)AI development for clinical practice. Application directions that are mentioned more frequently are lowering the administrative burden for medical personnel, forecasting complications and estimating potential risks and risk prevention, early warning systems, and tools that help monitor patients while they are out-of-hospital, for example via wearable devices.

\paragraph{Development of visual explanations for AI models.}
In terms of ease-of-use, visual explanations are indicated to be relatively easy to interpret as they remove potential language barriers. This means there is potential for research focused on the development of visual explanations in XAI.

\subsection{Requirements for successful implementation of (X)AI in practice}

\paragraph{More emphasis on multidisciplinary projects is needed for the development of (X)AI techniques.} 
While there is a large body of research on (X)AI-based tools for clinical practice, only very few make it to the bedside. Many reasons for this slow implementation can be attributed to a mismatch between all the different parties involved. Multidisciplinary collaboration is key to building a successful product, and \citet{van2022developing} propose a structured approach to develop and safely implement AI in medicine.

\paragraph{(X)AI training is required for clinicians.} 
As it stands momentarily, medical education in the Netherlands does not include obligatory classes on AI or AI-based medical devices, which can have significant consequences. First, understanding the background of the model, such as the training data used and how it was validated, is important to judge its performance and get a sense of trust. 
Second, it is often mistaken that explanations imply causality. Using AI-based tools without sufficient background knowledge makes the users prone to misinterpret the information provided. Third, some explanations may be too simplistic and may give a false sense of understanding, like feature importance plots. Instead, explanations should stimulate critical thinking so that the user can evaluate the given output themselves. Providing too simplistic explanations may hide crucial pieces of information that otherwise would raise doubts. Especially when the provided information is in line with the user's expectations, simple explanations may increase the risk of confirmation bias. Confirmation bias has indeed been shown to exist in human-technology interaction \cite[\textit{e.g.,}][]{wan2022bias,Bauer2023_XAI} and has been acknowledged in the XAI community as well \cite[\textit{e.g.,}][]{Ghassemi.2021, rudin2019_stopexplaining, cina2023semanticmatch}. On the other hand, experienced users may disregard a simple explanation altogether as they may find it unbelievable because the true reason for a given output is likely not as simple as a feature importance plot might suggest. After all, why are we using AI if it really were this easy?

%

Besides, without proper training, we might run the risk of two groups forming among clinicians: those who readily accept and trust AI-based tools, and those who are too resistant to use it. The first group might simply not be aware of important background information that could impact the usage and interpretation of the output. The second group might be too skeptical and hesitant to use technology they do not understand. Both could end up in effects opposite to what we want to achieve with the implementation of AI-based DSTs in clinical practice. Hence, training is key to a successful implementation of these tools as acknowledged in the previous work of \cite{tjoa2021_review}.

\paragraph{(X)AI tools should include more than just explanations.}
It is apparent that clinicians desire more than a simple explanation of the prediction, as many have pointed out that they would like to see a certainty score, which is also in line with observations of \cite{pmlr-v106-tonekaboni19a}. Additionally, we can confirm findings of \cite{Cai2019, lee2024improvinghealthprofessionalsonboarding} that suggest that clinicians are interested in the background of the model, such as information about the training data or benchmark performances. Transparency on training data and representativeness are also important to build trust in the model. A comparison can be made with the setup of clinical trials and evidence-based medicine.

\paragraph{We need more awareness of change management and how to navigate it.}
The world is changing at a high pace and especially technological progress is faster than ever, but humans are known to resist change and adapt slowly. From a neurological perspective, humans have been shown to respond faster to familiar stimuli \citep{Yang2023-familiarity}, and hence, tend to stick to routines. In the workplace, this means that we tend to resist changing our routine or workflow that we are used to. Many barriers in the adoption of AI-based DSTs are psychological in nature, as one of the interviewee says (interview 8), and hence we need to draw on knowledge from that field to overcome these barriers. Hence, the development of products for clinical practice needs to be linked to research on human cognition and psychology to allow for a fast and smooth adoption of the technology. 

However, while we should make sure to incorporate research from other fields into the development of these products to aid their implementation, people also need to experience its practical benefits over time and need to get used to using them. 

\subsection{Warnings regarding (X)AI usage and development}

\paragraph{Black-box models should not be used if they are not a significant improvement over glass-box models.} AI is supposed to be the future, but should it be used everywhere? Lately, the field receives more and more criticism for being blindly applied to any project, just because it is AI and it looks good. However, there are many situations where a much simpler model can achieve the same level of performance as a complex black-box model \citep{Semenova2022_10.1145/3531146.3533232}. 
Especially when it comes to high stakes situations, one should first use an interpretable glass-box model which allows clinicians to understand the mechanism, and only resort to more complex AI models if they sufficiently outperform glass-box models. 

\paragraph{AI should not increase the distance between clinicians and patients.} The goal of AI-based DSTs is to enhance medical care, and not to act as middleman between clinicians and patients. Hence, if there is a disconnect between both parties due to the introduction of such a technology, then we have missed the goal. DSTs in clinical practice should be intuitive to use to the extent that you should fail to notice their existence in the workflow. This has been coined as `unremarkableness' by \citet{Yang2019-unremarkableness}, who studied the degree of unremarkableness of AI to fit into the clinical decision-making process. To achieve this, we need contextual awareness early on in the design and development process.

\section{Conclusion}

This study offers insights from practicing clinicians on the integration of (X)AI in healthcare. Through semi-structured interviews, clinicians highlighted the potential benefits and challenges of (X)AI in clinical settings. 
While AI can improve decision-making, concerns were raised about the reliance on complex (X)AI models, which may erode trust and create distance between clinicians and patients. It is essential to strike a balance between AI's capabilities and human judgment to uphold patient-centered care.

One crucial takeaway is the importance of seamlessly integrating (X)AI into clinical workflows to enhance decision-making without causing disruptions. Clinicians emphasized the need for AI literacy and stressed the significance of collaboration among AI developers, healthcare professionals, and clinicians for successful adoption.
The study also delved into clinician preferences for explanation methods, with feature importance scores being favored for their simplicity and ease of comprehension. However, counterfactual explanations and rule-based methods were valued for providing deeper insights in intricate cases. Visual explanations were deemed essential by clinicians as they enhance the intuitiveness and accessibility of AI outputs, particularly in time-sensitive situations. 
The choice of explanation method should align with the clinical context — feature importances may be suitable for quick decisions, while decision rules or counterfactual explanations are more appropriate for detailed analysis or audits.

In conclusion, this study advocates for transparent (X)AI systems that prioritize visual clarity, multidisciplinary collaboration, thoughtful selection of explanation methods, and integration as supportive tools rather than barriers in healthcare practice based on real-world clinician input. Future research directions include investigating the role of responsibility as a construct distinct from trust. Additionally, collaborative research combining cognitive sciences with XAI to investigate the role of (X)AI in decision making is important. 
The field would benefit from a framework outlining the design process of (X)AI-based DSTs that incorporates multiple stakeholder perspectives. This is needed to ensure that these products can indeed be implemented seamlessly into the workflow and address the needs of clinicians. 

While this study offers valuable insights into clinicians' perspectives on XAI in healthcare, it has several limitations, including its narrow geographical focus on the Netherlands, reliance on qualitative data, potential selection bias, and lack of patient input or practical implementation. We believe this study provides a strong groundwork for future research addressing these limitations.


\bibliography{main}

\newpage
\appendix
\section{Interview methodology details}\label{app:methodology-details}

To report our study design, methodology and results we follow the guidelines published in the QOREC checklist \citep{coreq}. For conciseness, we have included the main aspects in the main body of the paper and leave the remaining details to this appendix.

The interviews were conducted by the first author (female) of this paper, who pursued their PhD in Transparent Machine Learning at the University of Amsterdam at the time the study was conducted. The interviewer obtained a university degree in Psychology with explicit training and analysis of qualitative studies like this one. While one potential interviewee did not follow up to arrange the interview after expressing interest (unknown reasons), however none withdrew their consent or dropped out after the informed consent form had been signed and the interview scheduled. All interviews were conducted in English and all except for one were held online via Zoom or Microsoft Teams depending on the interviewee's preferences. The reason being that most participants were located elsewhere in the Netherlands and online meetings were deemed most convenient. There was no-one else besides the interviewer and the interviewee present during the interviews because all meetings were recorded and field notes were not necessary as accurate transcripts could be created later from the recordings. Participants were also not asked to check the transcripts, although they were reminded at the end of the interview that they could always reach out in case they wanted to retract something they had said or would like to add more information. The supplementary material was made available to interviewees beforehand however there was no strict instruction to study it before the interview. The interview guide with prompts and questions was not shared with the interviewees. It is believed that data saturation has been reached as aspects raised by interviewees became increasingly redundant and no new angles were discussed. Participants expressed interest in the final manuscript once published and will be send a copy of the paper. 
\newpage

\section{Additional interview material}\label{app:interview_material}


The material below has been used during interviews to illustrate and explain three types of XAI methods: feature importances, counterfactual explanations, and rule-based methods.

\vspace{1cm}
\centerline{{\textbf{Types of Explanations for Machine Learning}}}
\vspace{0.5cm}
\noindent
In machine learning, we distinguish between glass-box models vs. black-box models. Glass-box models are those that are inherently interpretable, \textit{i.e.}, we see the model and are able to grasp how it is making predictions. Black-box models are complex machine learning models that we are not able to interpret directly. For those models, we see the predictions and then apply methods that generate post-hoc explanations. In this study we are focusing on the most prevalent types of explanations used: weighted rules as glass-box model, and feature importances and counterfactual explanations as post-hoc explanations for black-box models.

\begin{figure}[h]
    \centering
    \includegraphics[width=\textwidth]{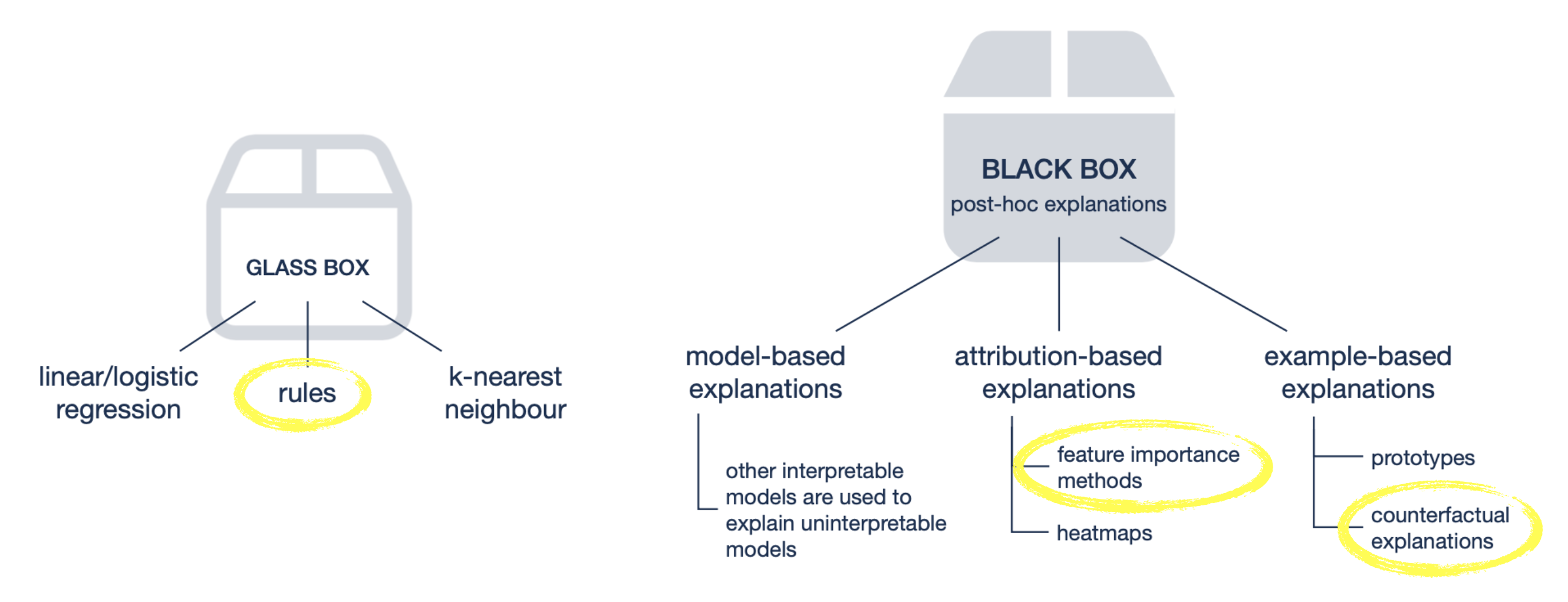}
    \caption{Overview of XAI methods}
\end{figure}

\noindent
\textbf{Toy example}\\
\\
\noindent For the sake of exemplifying the type of explanations, consider this toy example: We have a dataset consisting of 200 intensive care patients. For each patient, different medical parameters such as respiratory rate, oxygen saturation, systolic blood pressure, and so on, have been measured. We call such parameters features and they're named X1 to X5 in the dataset for simplicity. Please note that the values are intentionally chosen arbitrarily to keep the discussion general and avoid expert knowledge to factor in. These features are used to predict a binary (0 or 1) target variable like mortality, \textit{i.e.}, not-deceased (0) or deceased (1). A machine learning model has been fit to the data to predict the binary target variable. An excerpt of the data is shown below where the last column corresponds to the prediction of the model for the target variable. The highlighted row is the patient of interest, \textit{i.e.}, the patient for which the prediction is to be explained.

\begin{figure}[h!]
    \centering
    \includegraphics[width=0.35\textwidth]{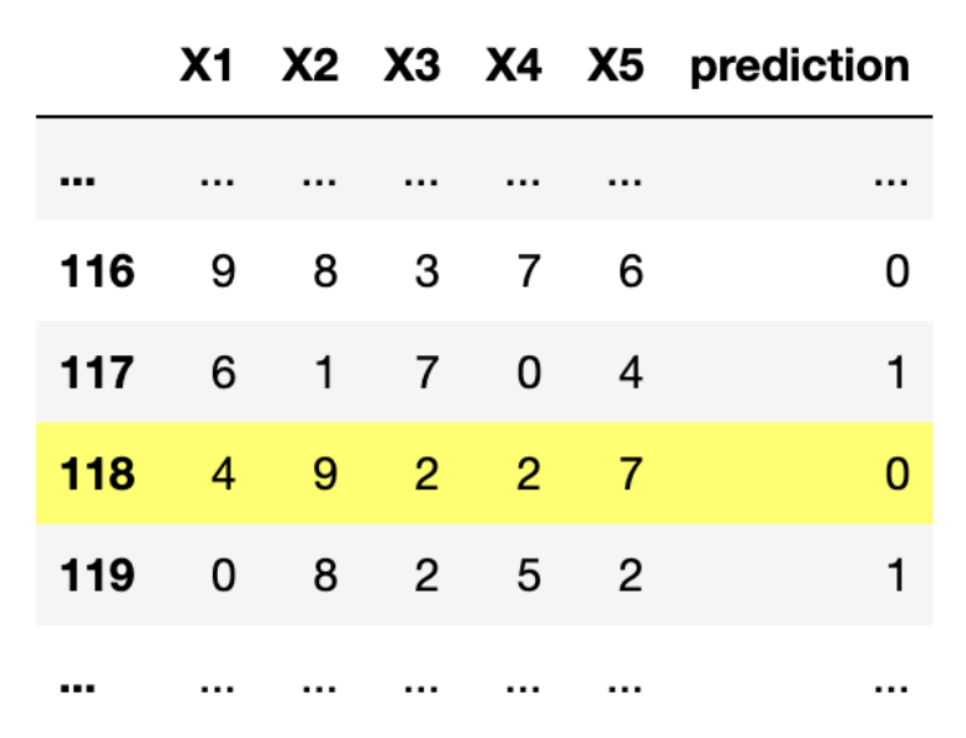}
    \caption{Excerpt of toy example data}
\end{figure}

\vspace{2cm}
\noindent
\textbf{Types of explanations}\\
\\
\noindent
\textit{Feature importances}\\
We see here the contribution of each feature to the average prediction visualised in a bar plot, for the highlighted patient in the example dataset. The prediction for that patient is 0 (not-deceased), and from the plot we can see that X5 contributed most to this prediction. The colors of the bars, red and blue respectively, indicate whether a contribution is positive (towards predicting 1) or negative (towards predicting 0). In this example, a positive contribution means that it is a risk increasing factor for death in this model, and a negative contribution indicates a risk decreasing factor. The length of the bar corresponds to the weight of the contribution. In this example, X5 has the biggest impact and is a risk decreasing factor.

\begin{figure}[h]
    \centering
    \includegraphics[width=0.6\linewidth]{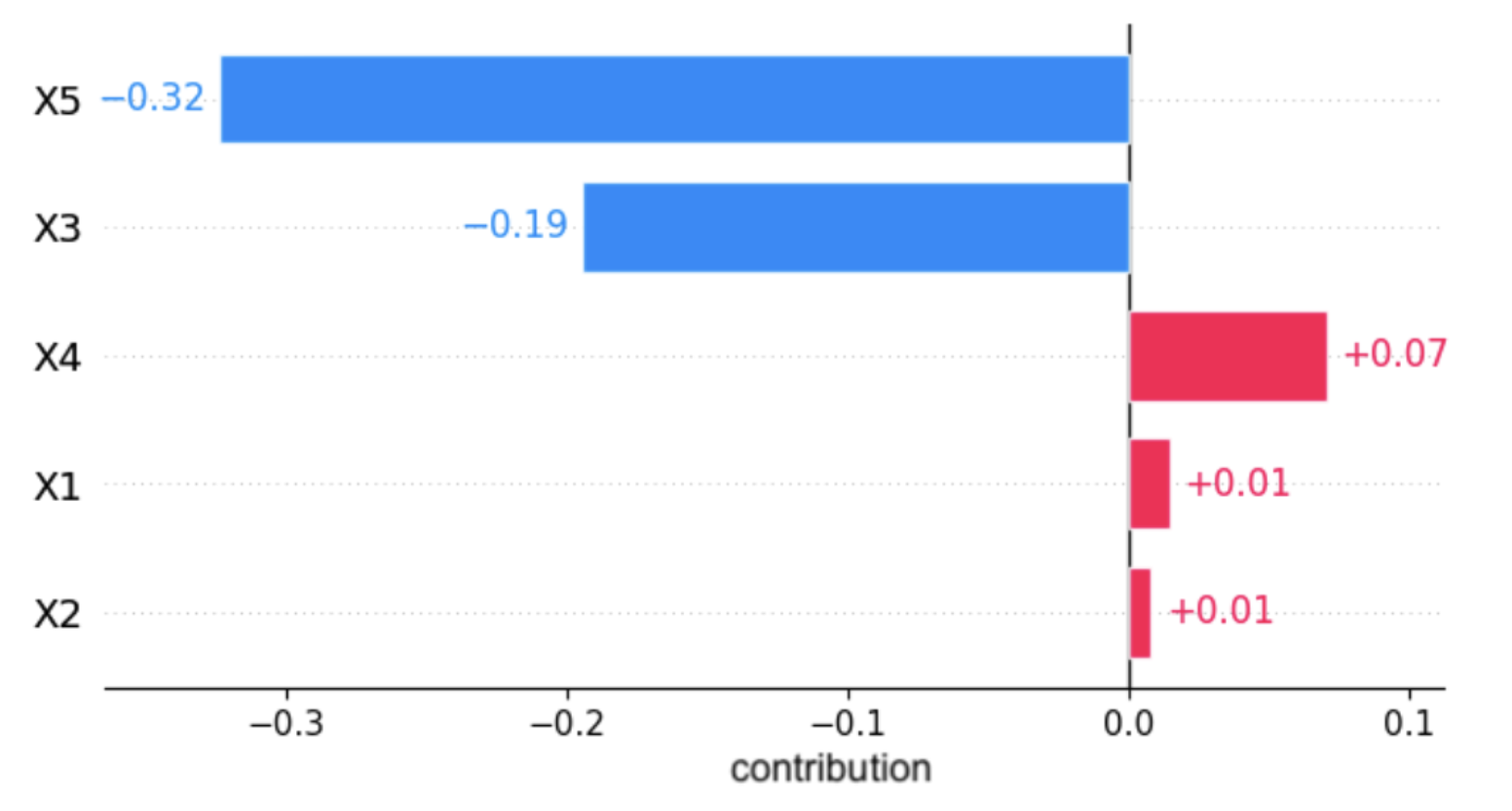}
    \caption{Feature importances (SHAP) example}
\end{figure}

\noindent
\textit{Counterfactual explanations}\\
Counterfactual explanations show how the feature values of a patient should change such that the prediction would also change. The instance of which you want to explain the prediction is called the \textit{factual}. In our case, the \textit{factual} is the highlighted patient in the toy example, for whom the prediction is 0 (not-deceased). A \textit{counterfactual} is another (possibly fictitious) patient, who is very similar to the \textit{factual} with regard to the feature values (say equal heart rate, systolic blood pressure, and respiratory rate, but slightly different temperature and oxygen saturation), who is however predicted as 1 (deceased).

\noindent In the table below, this is visualised. The \textit{factual} is the highlighted patient from our dataset that we want to derive an explanation for. The \textit{counterfactual} is another patient, with equal values for X2, X4, and X5 as compared to the \textit{factual}. However, for this other patient, X1 has a value of 2 and X3 has a value of 5. The prediction for this patient is 1 (deceased).

\begin{figure}[h]
    \centering
    \includegraphics[width=0.4\linewidth]{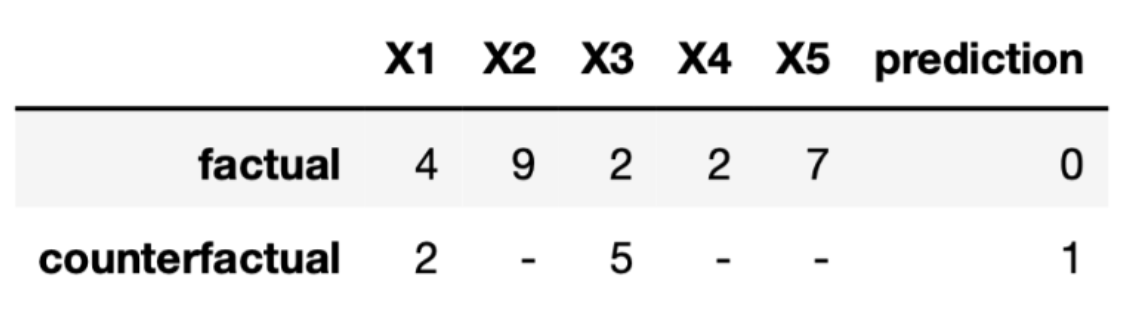}
    \caption{Counterfactual explanation example}
\end{figure}

\noindent
\textit{Weighted rules}\\
Instead of explaining a prediction made by a machine learning model, we can derive predictions ourselves using a set of rules. These rules follow the form: ``If ... AND ..., then PREDICT ... .''. For example,``IF systolic blood pressure is higher than 130 AND respiratory rate is higher than 25, then predict 1 (deceased''. If a rule does not apply to a patient, \textit{i.e.}, the IF-conditions are not met, this rule is not used for the patient. To predict the outcome for a patient, we use all rules for which the patient fulfils the IF-conditions. Each rule makes a prediction, which can be contradicting each other – some rules may predict 0 (not-deceased), some may predict 1 (deceased). Each rule comes with a weight, which can be thought of as the importance of the rules. For each class (not- deceased vs. deceased), we sum the weights of the respective rules. The final prediction will be the class with the largest sum of weights.\\
\\
For the toy example the algorithm returns four rules, visually displayed below.\\

\begin{figure}[h]
    \centering
    \label{fig:rug}
    \includegraphics[width=0.8\linewidth]{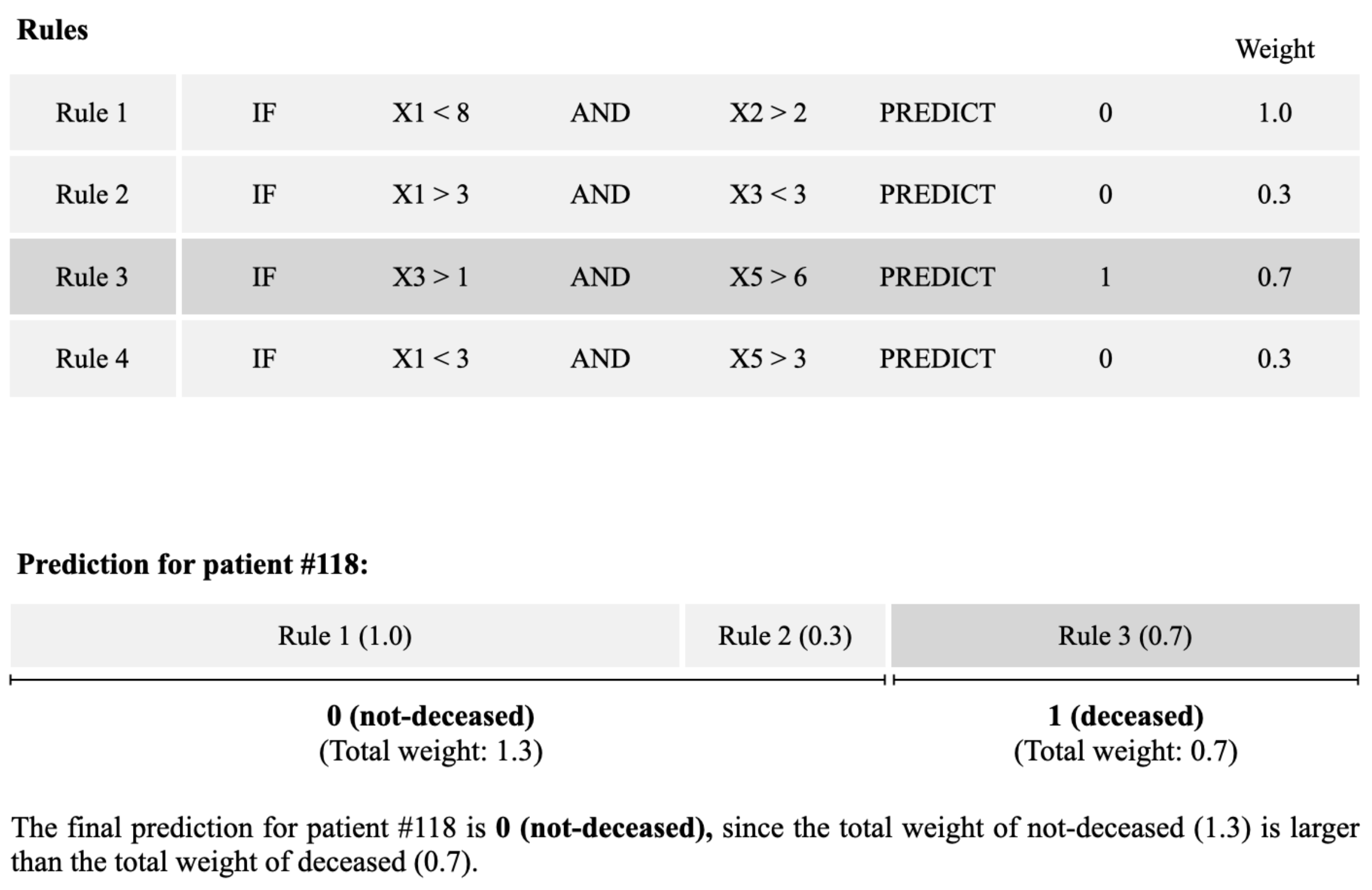}
    \caption{Weighted rules example}
\end{figure}

\noindent Using these rules, we can make the prediction for the highlighted patient: According to rule (1) and (2), the prediction for the highlighted patient would be class 0 (not-deceased). The summed weight for these rules is 1.0 + 0.3 = 1.3. According to rule (3), it would be predicted as class 1 (deceased), with a weight of 0.7. Rule (4) is not used for our patient, since the first condition of the rule (\textit{i.e.}, X1 < 3) is not met by the patient who has a value of X1 = 4. Hence, we use only the first three rules. To make the final prediction, we look at the summed weights: 1.3 and 0.7 for class 0 (not-deceased) and 1 (deceased), respectively. Hence the final prediction for the patient is 0 (not- deceased).

\end{document}

%% file: tables/interview-scheme.tex
{
\begin{table}[ht!]
\centering
\caption{Interview scheme}
\label{tab:interview}
\resizebox{\textwidth}{!}{%
\begin{tabular}{@{}p{0.3\textwidth}p{0.8\textwidth}@{}}
\hline\hline
\textbf{Topic}  & \textbf{Questions}      \\
\hline
\rowcolor{gray!10} \multicolumn{2}{l}{\textbf{Opening}} \\
(a) Establishing rapport & [interviewer thanks participant for their time; interviewer introduces themselves] \\
(b) Purpose of the study & [interviewer shortly introduces XAI and explains purpose of the study] \\
(c) Procedure & [interview will last about 30 minutes, an audio recording will be started] \\
\hline
\rowcolor{gray!10} \multicolumn{2}{l}{\textbf{Body}} \\
(a) Background &  Could you tell me a bit more about field of work and your expertise? \\
& What specialization do you work in? \\
& How long have you been working in this field? \\
(b) AI in healthcare & In healthcare generally, and also in your field, are AI systems in use?  \\
& Could you describe the extent to which you are confronted with automatic systems that use AI in your daily work? \\
& For what type of tasks are AI systems used? \\
& Can you tell me more about your research on AI for [...]? \\
& Can you think of AI systems that are not (yet) in use but that you could imagine helpful? \\
& What do you believe are the reasons why implementation of AI in practice is slow? \\
(c) Attitude towards AI & Could you describe your positive and negative feelings about using AI in healthcare practice? \\
& What are your thoughts on trust in AI? \\
(d) Explanation methods & [interviewer gives overview and example of explanation methods using a toy example]\\
& For each of these methods, can you briefly describe how familiar you are with them?  \\
& What are your thoughts on each of these methods? \\
& What are desirable characteristics of an explanation for an AI system? \\
\hline
\rowcolor{gray!10} \multicolumn{2}{l}{\textbf{Closing}} \\
(a) Final comments & Is there anything else that you would like to share with me on this topic? \\
& Do you have any questions to me? \\
(b) Closing & [interviewer thanks interviewee again for their time and participation]\\
\hline\hline
\end{tabular}%
}
\end{table}
}